\documentclass[11pt]{article}

\usepackage[final]{acl}

\usepackage{times}
\usepackage{latexsym}
\usepackage{amsmath}
\usepackage[most]{tcolorbox}
\usepackage{tabularx}
\usepackage{booktabs}
\usepackage{placeins}
\usepackage{fontawesome5}
\usepackage[T1]{fontenc}
\usepackage[utf8]{inputenc}
\usepackage{url}
\usepackage{microtype}

\usepackage{inconsolata}

\usepackage{graphicx}

\title{Rating the Raters: Rasch Measurement Theory for LLM Evaluation}

\author{
Pratik S. Sachdeva \\
University of California, Berkeley \\
\texttt{pratik.sachdeva@berkeley.edu}
\And
Nathan Boudol \\
Grenoble INP \\
\texttt{nathan.boudol@grenoble-inp.org}
\AND
{\normalfont
\href{https://github.com/pssachdeva/rating_the_raters}
     {\faGithub\ Code}
\quad
\href{https://huggingface.co/datasets/psachdeva/rating-the-raters}
     {\faDatabase\ Data}
}
}

\begin{document}
\maketitle
\begin{abstract}
LLMs now sit on every side of evaluation: as \textit{examinees} scored on benchmarks, \textit{judges} of other models' outputs, and \textit{raters} of human-generated content. Each paradigm can be viewed as a \textit{measurement} problem, where a latent property of an object is probed with items from an instrument (e.g., benchmark) by judges or raters. Standard evaluation practices often neglect the contributions of each core component to the end result, limiting our understanding of what is being measured. Rasch measurement theory (RMT) is well-suited to this problem. RMT decomposes ordinal ratings into separable \textit{facets} on a common scale. It further provides a battery of diagnostics that can identify miscalibrated measurements and rater biases. We present a case study of RMT applied to the LLM-as-rater paradigm using the \textit{Measuring Hate Speech} corpus, whose construct was itself built under RMT. We fit a series of many-facet Rasch models to annotations from nine LLMs spanning families and capability levels. Our analyses show that LLMs systematically differ from human raters in severity, item-level calibration, question-order robustness, target-identity sensitivity, and rating scale use, all of which standard evaluation practice would largely obscure. Overall, we argue that RMT belongs in the toolkit for evaluating LLM-as-examinee, -judge, and -rater paradigms.
\end{abstract}

\section{Introduction}
LLMs now sit on every side of evaluation. As \textit{examinees}, they are scored on benchmarks that estimate their capabilities and character traits \cite{hendrycks2021measuring, srivastava2023beyond}. As \textit{judges}, they evaluate the outputs of other LLMs against rubrics \cite{zheng2023judging, li2024llms}. As \textit{raters}, they label human-generated text for specific properties, like sentiment or hate speech \cite{gilardi2023chatgpt, calderon2025alternative}. These paradigms, at their heart, reflect \textit{measurement} problems. 

From a measurement perspective, an evaluation requires defining a \textit{construct}, or the latent property -- a model's reasoning ability, an LLM's helpfulness, or a social media comment's hatefulness -- we aim to measure. We quantify the construct as a score by developing a measurement \textit{instrument} containing \textit{items} (or tasks) designed to probe the object of measurement at various points of the construct. The responses on each item are tabulated into a final score, thereby measuring the latent property. If the instrument is not adequately designed, the measurement fails, and its scores cannot be trusted to reflect the construct \cite{Wallach2025GenAIMeasurement, Weidinger2025EvaluationScience, Zhou2026GeneralScales}. For example, a software engineering benchmark whose items are overly simplistic or can be gamed via reward hacking lacks construct validity, since it does not adequately measure engineering ability.

Thoughtful measurement design is particularly important as evaluations move toward subjective tasks where the LLM may act as examinee, judge, or rater. LLM-as-judge and LLM-as-rater often rely on subjective rubrics where respondents can reasonably disagree \cite{basile2021need, plank2022problem}. These constructs are typically validated with annotator agreement metrics on ``gold labels'' generated by humans \cite{rottger2022two}. These reporting practices, however, may obscure problematic measurement by collapsing different components of the measurement (the examinee, the items, and the rater) into a single score \cite{Salaudeen2025MeasurementToMeaning}. More critically, this superficial treatment avoids answering a deep question: \textit{who} sets the measurement scale -- via the construct and constituent item design -- when designing evaluations? If humans set the measurement scale, can we be sure that the same scale translates neatly to LLMs?

\textit{Rasch measurement theory} (RMT) addresses both concerns \cite{rasch1993probabilistic, andrich1988rasch, linacre1989manyfacet}. RMT decomposes ordinal ratings (e.g., Likert-style rubrics) into separable contributions from the object being measured (here, a social media comment), the items (the tasks), and the raters (either LLMs or humans), all placed on a common measurement scale. It is similar to (and often overlaps with) item response theory (IRT) and psychometrics, which have previously been used in evaluation for LLM-as-examinee settings~\cite{lalor2016building, vania2021comparing, Federiakin2025Leaderboards, Chen2025CompactIRT, Yao2025JEIRT, Ye2025LLMPsychometrics} and more recently in the LLM-as-judge and -rater settings~\cite{giorgi2025human, Choi2026JudgeIRT, Feng2026NoisyJudges, Chen2026NoisyJudgeInference, Jiao2026RaterEffects}. However, RMT is distinct in its commitment to two additional requirements: \textit{invariance}, which requires that the instrument function the same way across raters and items, and a principle that the evaluation data must fit a fixed measurement model rather than adjusting the model to the data. Furthermore, RMT comes equipped with a battery of diagnostics allowing practitioners to interrogate the resulting measurement scale, determining whether it is suitably calibrated for the construct at hand.

In this work, we present a case study of RMT applied to the LLM-as-rater paradigm, focusing on hate speech measurement. We use this domain because (i) hate speech is highly subjective and relevant for LLM-as-rater, and (ii) the \textit{Measuring Hate Speech} (MHS) corpus -- 50{,}070 comments labeled by 11{,}143 annotators on a ten-item instrument -- was itself constructed under RMT, giving us a validated construct against which to compare LLM annotations \cite{DBLP:journals/corr/abs-2009-10277, sachdeva-etal-2022-measuring}. We collect annotations from nine LLMs spanning families, providers, and capability levels, and fit a series of many-facet Rasch measurement models in which the measurement scale is calibrated by humans alone, by LLMs alone, or anchored to the human scale. Our analyses seek to answer the following questions:
\begin{itemize}
\item \textbf{RQ1:} Can LLM annotations calibrate a measurement scale using a construct developed and validated only with human annotations? How does the resulting scale compare with a human-calibrated scale? (Section~\ref{sec:rq1})
\item \textbf{RQ2:} Given a human-calibrated measurement scale, how do LLM raters compare to human raters? (Section~\ref{sec:rq2})
\item \textbf{RQ3:} How robust are LLM ratings to specific survey items, question order, and the target identity described by a comment? (Section~\ref{sec:rq3})
\end{itemize}
Our goal is methodological. We offer a case study of how RMT supplies a principled vocabulary for interrogating LLM-as-raters, and we argue that the same vocabulary belongs in the toolkit used to evaluate LLMs across the examinee, judge, and rater paradigms. Across our analyses, we find that LLMs depart from human raters in structured and interpretable ways, which would be harder to detect without RMT. We present an overview of RMT in Section~\ref{sec:rmt}, the methods and data we use in Section~\ref{sec:methods}, and results in Section~\ref{sec:results}.

\section{Rasch Measurement Theory}
\label{sec:rmt}
The goal of measurement theory is to make observations of some property meaningfully comparable. In measurement settings, observations are typically produced by the interaction of three components: the objects whose latent property is being measured, the items of a survey instrument that probe that property, and the raters who use the instrument to assign ordinal responses. Specific measurement frameworks allow one to transform these observations into variables that reflect an underlying scale, facilitating cross-rater comparisons. Item response theory (IRT) is one approach that utilizes probabilistic models to extract continuous scales from ordinal responses such as Likert ratings. 

Rasch measurement theory (RMT) is a distinct methodological tradition that commits to \textit{invariance} \cite{rasch1993probabilistic, andrich1988rasch}. Invariance requires that the measurement instrument function the same way regardless of which raters produce the responses, which objects are rated, and which items are used to construct the scale. RMT achieves invariance in two ways. First, it uses a specific class of additive IRT models -- here, we focus on the \textit{many-facet model} -- that mathematically guarantees invariance \cite{linacre1989manyfacet}. Second, it requires that the observational data fit the model rather than the model fit the data \cite{bond2015applying}. Thus, a poor fit should be read as information about the items, the raters, or the construct, but not as a problem to be solved by adjusting the model.

\textbf{Construct Theorization.} Constructing a Rasch measurement scale begins with theorizing a \textit{construct}: an explicit articulation of the latent property being measured and how it varies along a continuum \cite{Wallach2025GenAIMeasurement}. For hate speech, the construct specifies what it means for a comment to lie at successive points along the spectrum: from neutral or supportive speech, through expressions of disrespect, insult, humiliation, and dehumanization, to incitement of violence and calls for genocide. Developing a construct requires careful qualitative analysis of constituent examples; it is operationalized via a \textit{survey instrument} whose items map onto distinct regions of the spectrum. We adopt the construct established by prior work on the \textit{Measuring Hate Speech} corpus, whose survey instrument contains ten items discussed further in Section~\ref{sec:corpus} \cite{DBLP:journals/corr/abs-2009-10277, sachdeva-etal-2022-measuring}.

\textbf{Many-Facet Rasch Model.} The many-facet model recovers a continuous scale from the ordinal responses on the ten items of the MHS instrument. It captures the decision to opt for response $k$ (say, ``strongly agree'') versus response $k-1$ (``agree''). Let $p_{nijk}$ be the probability that rater $j$ assigns comment $n$ a response $k$ on survey item $i$; similarly define $p_{nij(k-1)}$ for response $k-1$. The model defines an odds $r_{nijk}$ as:
\begin{equation}
    \log r_{nijk} = \log \frac{p_{nijk}}{p_{nij(k-1)}} = \theta_n - \delta_i - \alpha_j - \tau_{k}. \label{eqn:base_irt}
\end{equation}
Items are coded so that higher $k$ always corresponds to greater hatefulness. A larger log-odds, therefore, indicates that the rater is more inclined to assign the more-hateful category. The log-odds depends on four additively separable terms:
\begin{itemize}
    \item $\theta_n$, the \textbf{hate speech score} of comment $n$. Higher $\theta_n$ indicates a more inherently hateful comment. 
    \item $\delta_i$, the \textbf{difficulty} of survey item $i$. Difficulties provide a check on construct validity: we should expect the fitted values to follow the ordering of the construct.
    \item $\alpha_j$, the \textbf{severity} of rater $j$. Raters with higher severity are less likely to label comments as possessing features of hate speech: their thresholds for ``hatefulness'' are higher.
    \item $\tau_k$, the \textbf{Rasch-Andrich threshold} for response $k$, whose ordering indicates whether the ordinal responses function appropriately.
\end{itemize}
Because comments, items, and raters share a common logit scale, model parameters can be interpreted together in the context of the construct.

\section{Methods}
\label{sec:methods}
\subsection{The \textit{Measuring Hate Speech} Corpus}
\label{sec:corpus}
The \textit{Measuring Hate Speech} corpus consists of 50,070 social media comments labeled by 11,143 human annotators. Annotators rated each comment using a survey instrument containing ten items spanning a spectrum of potential hatefulness: \textit{Sentiment}, \textit{Respect}, \textit{Insult}, \textit{Humiliate}, \textit{Status}, \textit{Attack/Defend}, \textit{Dehumanize}, \textit{Violence}, \textit{Genocide}, and an overall \textit{Hate Speech} item. Each item consists of five Likert-like responses, other than \textit{Hate Speech}, which is ternary. The corpus also contains annotations denoting the identity group(s) targeted by each comment. We focus on identity categories spanning gender, race/ethnicity, religion, and sexuality. 

We obtained the MHS corpus from Hugging Face in compliance with the authors' license and intended use \cite{ucberkeleyDlabMHS}. We consider two subsets of the MHS dataset for obtaining LLM ratings: (i) the \textbf{reference comment set}, consisting of 70 comments, each with more than 200 annotations, that served as the foundation for the original measurement scale, and (ii) the \textbf{expanded comment set}, consisting of 5,990 comments, each annotated by at least four human raters, of whom at least 75\% agreed on a single target identity.

\begin{table*}[t]
\centering
\small
\begin{tabularx}{\textwidth}{@{}llXrll@{}}
\toprule
Section & Figure & Analysis & Equation & Comment set & Model set \\
\midrule
\ref{sec:rq1} & Fig.~\ref{fig:wright_map} & Comparison of LLM- and human-calibrated measurement scales
    & Eq.~\ref{eqn:llm_scale} & Expanded comment set & Primary model set \\
\ref{sec:rq1} & Fig.~\ref{fig:rasch_andrich_thres} & LLM rating-scale functioning
    & Eq.~\ref{eqn:llm_scale} & Expanded comment set & Primary model set \\
\ref{sec:rq2} & Fig.~\ref{fig:model_severities}a & LLM severity relative to human raters
    & Eq.~\ref{eqn:model1} & Reference comment set & Primary model set \\
\ref{sec:rq2} & Fig.~\ref{fig:model_severities}b & Association between LLM severity and AA Intelligence Index
    & Eq.~\ref{eqn:model1} & Reference comment set & Full model set \\
\ref{sec:rq3} & Fig.~\ref{fig:severity_decomposition} & Item-level calibration
    & Eq.~\ref{eqn:model2} & Reference comment set & Primary model set \\
\ref{sec:rq3} & Fig.~\ref{fig:item_order} & Question-order differential rater functioning
    & Eq.~\ref{eqn:model1} & Reference comment set & Primary model set \\
\ref{sec:rq3} & Fig.~\ref{fig:target_identity_drf} & Target-identity differential rater functioning
    & Eq.~\ref{eqn:model3} & Expanded comment set & Primary model set \\
\bottomrule
\end{tabularx}
\caption{Summary of all analyses presented in the main text, along with corresponding sections, RMT models, figures, comment sets, and model sets.}
\label{tab:analysis-summary}
\end{table*}

\subsection{Language Models}
\label{sec:models}
We consider two distinct sets of models in our analysis. First, the \textbf{primary model set} consists of nine LLMs spanning families, providers, and capability levels: Claude Opus 4.6, Gemini 3.1 Pro, GPT-5.4, Grok 4.1, DeepSeek V4 Pro, Kimi K2.5, MiniMax M2.5, GPT-OSS 120B, and Llama 3.3 70B. We queried the primary model set with both the reference comment set and the expanded comment set. Second, the \textbf{full model set} consists of 33 models for which we queried only the reference comment set. See Appendix~\ref{sec:all_llms} for tables of the primary model set and the full model set with additional metadata. We prompted each LLM with a system prompt that described the task, provided the survey items verbatim from the MHS instrument, and specified a JSON response format (see Appendix~\ref{sec:prompts} for prompts). We queried each LLM (default temperature of 1, where applicable) via batch or asynchronous APIs. Closed models were accessed through their provider, and open-weight models through either their developer's API or third-party inference providers (OpenRouter, Together AI). We used ``medium'' reasoning for all models, where applicable.

\subsection{Fitting Measurement Models}
\label{sec:measurement_fit}
We fit three classes of measurement models: (i) human-only, (ii) human-anchored, and (iii) LLM-only. The human-only measurement model used the full MHS corpus with a recoding scheme that improved the calibration of the measurement scale: \textit{Insult} and \textit{Attack/Defend} were collapsed to four responses, \textit{Humiliate} to three, and \textit{Status}, \textit{Dehumanize}, \textit{Violence}, \textit{Genocide}, and the ternary \textit{Hate Speech} item were binarized. The human-anchored models had hate speech scores $\theta_n$ and item difficulties $\delta_i$ anchored to those of the human-only measurement scale, with the remaining parameters fit to LLM annotations. The LLM-only measurement model was fit to LLM annotations with no anchoring. We fit all measurement models using the \textit{Facets} software, version 4.5.0 \cite{Linacre2026Facets}.

\section{Results}
\label{sec:results}

We organize the results around our three research questions. Section~\ref{sec:rq1} asks whether LLM annotations can calibrate a measurement scale and compares the resulting scale with its human-calibrated counterpart. Section~\ref{sec:rq2} places LLM raters on the human-calibrated scale. Section~\ref{sec:rq3} examines robustness to specific survey items, question order, and target identity. Table~\ref{tab:analysis-summary} maps analyses to their respective measurement model, comment set, and model set. See Appendix~\ref{sec:prompts} for system prompts and Appendix~\ref{sec:stats} for the results of all statistical tests.

\begin{figure*}[t]
    \centering
    \includegraphics[width=0.83\linewidth]{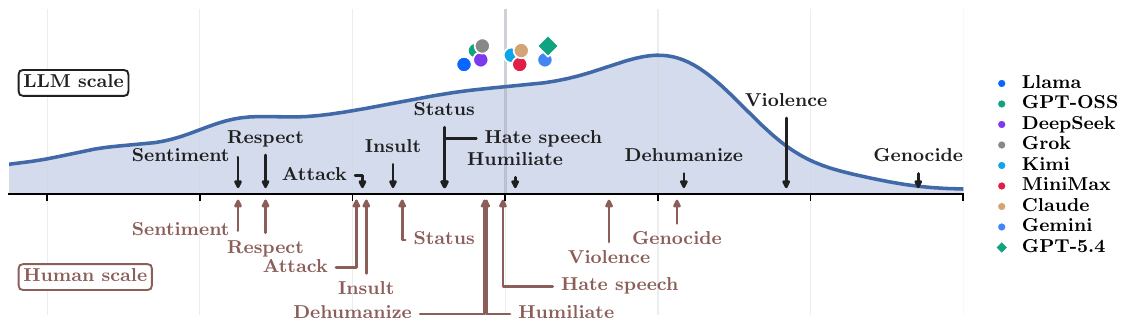}
    \caption {\textbf{Comparison of LLM and Human Measurement Scales.} Horizontal axis denotes measurement scale for all figure elements. Distribution of comment scores for LLM scale denoted by blue curve. Item markers on top half (black) refer to LLM difficulties; item markers on bottom half (brown) refer to human difficulties linked to the LLM scale via \textit{Sentiment}. Model severities denoted by colored points; jitter added for separability. Legend orders models according to increasing severity. Thicker vertical gray line denotes zero-line for measurement scale.}
    \label{fig:wright_map}
\end{figure*}

\subsection{Constructing an LLM-Only Scale}
\label{sec:rq1}
In many deployment settings, LLM judges or raters will set the measurement scale themselves rather than being placed on a human-calibrated scale. This raises important questions about whether construct validity and the survey instrument -- often designed with humans in mind -- translate well to the LLM context. Any divergence between LLM- and human-derived scales would mean that the resulting scores carry different meanings than the instrument was designed to measure. Thus, we asked: \textit{Can LLM annotations calibrate a measurement scale using a construct developed and validated only with human annotations?}

The MHS corpus is well-suited for comparing calibrated LLM and human measurement scales on the same construct. We fit the measurement model
\begin{equation}
\log r_{nijk} = \theta_n - \delta_i - \alpha_j -\tau_k \label{eqn:llm_scale}
\end{equation}
where all of $(\theta_n, \delta_i, \alpha_j, \tau_k)$ are free parameters. We used annotations from the primary model set on the expanded comment set from the MHS corpus (up to 53,910 samples total). We then examined the category functioning ($\tau_k$ values) and compared the LLM and human measurement scales.

% For example, in the MHS corpus, the \textit{Genocide} survey item asked raters whether they agreed that the comment called for genocide of a target identity group using a 5-point Likert scale (``strongly disagree'' to ``strongly agree''). The response thresholds -- $\tau_k$ in Equation~\ref{eqn:llm_scale} -- were found to be miscalibrated, suggesting that human raters were not using the five responses consistently. This motivated collapsing the rating scale from a 5-point Likert to a binary scale: effectively, does the comment call for genocide or not?

\textbf{Comparing the LLM and Human Measurement Scales.} We first compare the LLM measurement scale with the human measurement scale. To simplify the comparison, we fit Equation~\ref{eqn:llm_scale} to the LLM annotations using the human recoding scheme to ensure both scales operate over identical response categories (see Section~\ref{sec:measurement_fit}) and extracted the comment scores $\theta_n$, item difficulties $\delta_i$, and the nine model severities $\alpha_j$. In general, comparing two measurement scales is not straightforward. We cannot, for example, simply compare the item difficulties between the two scales directly. We can, however, compare \textit{differences} on the logit scale: the difference between, e.g., the \textit{Sentiment} and \textit{Respect} items is comparable between the human and LLM scales. We thus applied a \textit{single-anchor additive linking} constant \cite{Slinde_1979}: we shifted each $\delta_i^H$ by a constant $\Delta = \delta_1^H - \delta_1^{\text{LLM}} $ where $\delta_1$ is the item difficulty for \textit{Sentiment}. We compare the LLM-scale difficulties, (adjusted) human-scale difficulties, the comment scores, and the LLM severities in Figure~\ref{fig:wright_map}.

We find that the lower end of the human construct -- \textit{Respect}, \textit{Attack}, and \textit{Insult} -- shows similar relative spacing across the human and LLM scales. This pattern diverges on the upper end of the construct: \textit{Dehumanize}, \textit{Violence}, and \textit{Genocide} exhibit much greater relative separation, suggesting that LLMs have a much higher threshold for these items. Two items also appear in different rank positions across the two scales: \textit{Hate Speech} occupies a lower-difficulty position on the LLM scale, while \textit{Dehumanize} occupies a higher one. The remaining items preserve their human-scale order. Despite differences in the placement of specific items, the hate speech scores obtained from each measurement scale exhibit high agreement in their order (Spearman correlation $\rho = 0.916$). Together, these results show that the LLM-only scale preserves the ordering of comments on the construct but diverges from the human scale at the upper end, where LLMs require greater endorsement to register severe items -- a pattern we examine further with item-level calibration diagnostics in Section~\ref{sec:rq3}.

\begin{figure}[t]
    \centering
    \includegraphics[width=\linewidth]{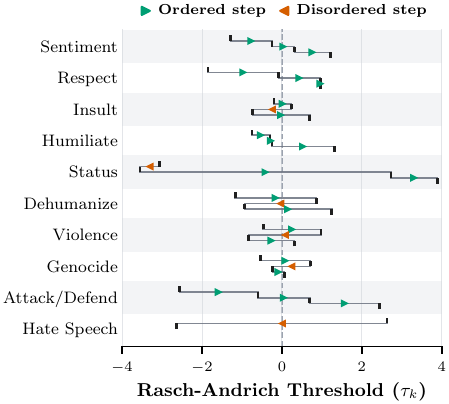}
    \caption {\textbf{Category functioning to assess response scale use.} The Rasch-Andrich thresholds $\tau_k$ ($x$-axis) obtained from the LLM measurement scale for each item ($y$-axis). Thresholds are denoted by vertical lines. Ordered steps, where $\tau_k > \tau_{k-1}$, are denoted by green triangles pointing to the right; disordered steps ($\tau_k < \tau_{k-1}$) are denoted by red triangles pointing to the left.}
    \label{fig:rasch_andrich_thres}
\end{figure}

\begin{figure*}[t]
    \centering
    \includegraphics[width=0.95\linewidth]{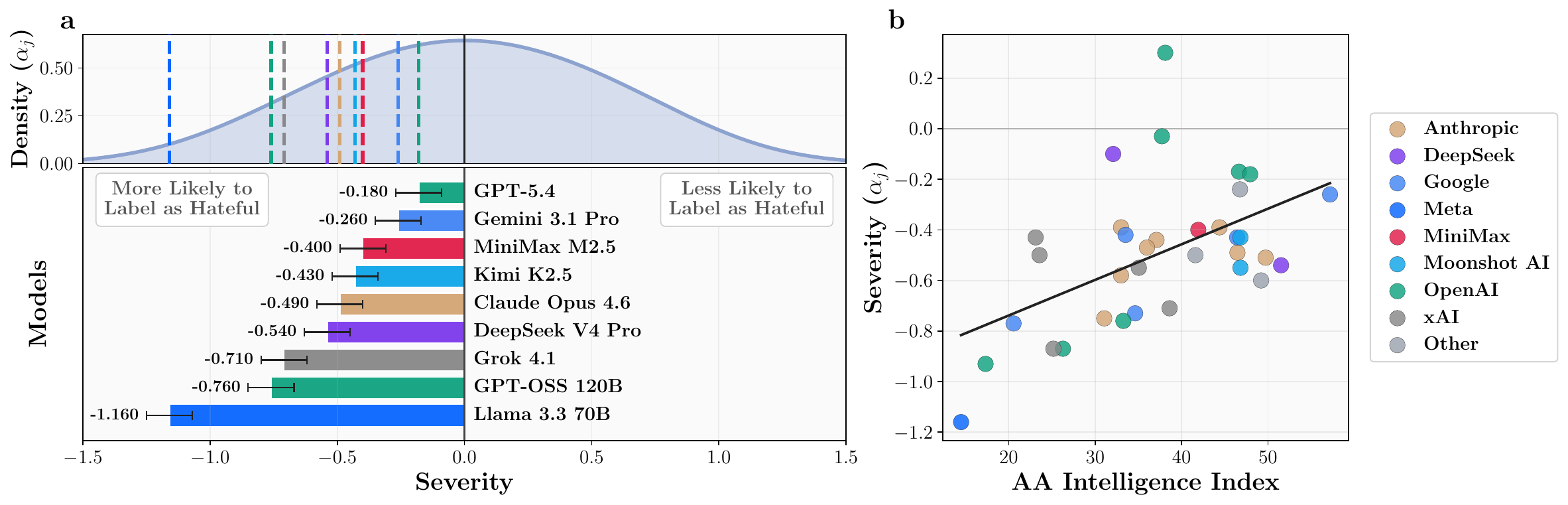}
    \caption {\textbf{Measurement models quantify LLM rater thresholds.} \textbf{a.} Fitted severity parameters $\alpha_j^{\text{LLM}}$ for nine LLMs (horizontal bars; error bars indicate standard error). Top plot compares parameter estimates (dashed colored lines) to the distribution of human severity parameters (blue curve: kernel density estimate). \textbf{b.} Comparison of severity estimates ($y$-axis) to AA Intelligence Index ($x$-axis) across the full model set (Pearson $r=0.52$, $p<10^{-3}$). Black line denotes linear fit. Bar and point colors denote model provider (legend shown panel \textbf{b}). }
    \label{fig:model_severities}
\end{figure*}

\textbf{Diagnosing Disordered Response Thresholds with Rating Scale Functioning.} An often overlooked consideration in evaluations is whether the rating scales associated with each item -- often Likert-like -- are well-constructed. For example, the \textit{Genocide} item in MHS originally had five responses, but was binarized by the corpus authors after the response thresholds were found to be miscalibrated for human raters. Since LLMs have been shown to exhibit systematic biases in their use of Likert rating scales~\cite{tjuatja2024llms, salecha2024llms}, it is worth leveraging RMT to examine whether rating scales translate to the LLM context.

We thus assessed the \textit{rating scale functioning} of each item via the $\tau_k$ thresholds, often referred to as \textit{Rasch-Andrich thresholds} \cite{Andrich1978Rating, Linacre2002Optimizing}. Specifically, each item has five responses (other than the \textit{Hate Speech} item, which has three), corresponding to four (or two) thresholds. A well-calibrated rating scale would always satisfy $\tau_k > \tau_{k-1}$: ``strongly agree'' should have a higher bar than ``agree'' in the log-odds. We plot the Rasch-Andrich thresholds for each item in Figure~\ref{fig:rasch_andrich_thres}. We mark ordered steps (where $\tau_k > \tau_{k-1}$) with green arrows pointing right, and disordered steps (where $\tau_k < \tau_{k-1}$) with red arrows pointing left. A well-calibrated scale, therefore, would look like three ordered steps pointing to the right. We find that some items -- like \textit{Sentiment}, \textit{Respect}, \textit{Humiliate}, and \textit{Attack} -- are well-calibrated. Other rating scales are slightly miscalibrated -- \textit{Insult} and \textit{Status}, for example, only have one disordered step, and thus would likely become calibrated after collapsing only one or two adjacent responses. The other three items -- \textit{Dehumanize}, \textit{Violence}, and \textit{Genocide} -- have $\tau_k$ values that generally do not exhibit any global ordering, suggesting they need stronger interventions. Notably, despite these disordered thresholds, Krippendorff's alpha indicated that most items exhibited high agreement among LLM raters (e.g., \textit{Violence} with $\alpha = 0.758$), suggesting that agreement alone cannot establish a functioning rating scale (Appendix~\ref{sec:agreement}). Thus, rating scale functioning can identify improperly calibrated ordinal levels in an instrument and guide their remediation.

\begin{figure}[t]
    \centering
    \includegraphics[width=\linewidth]{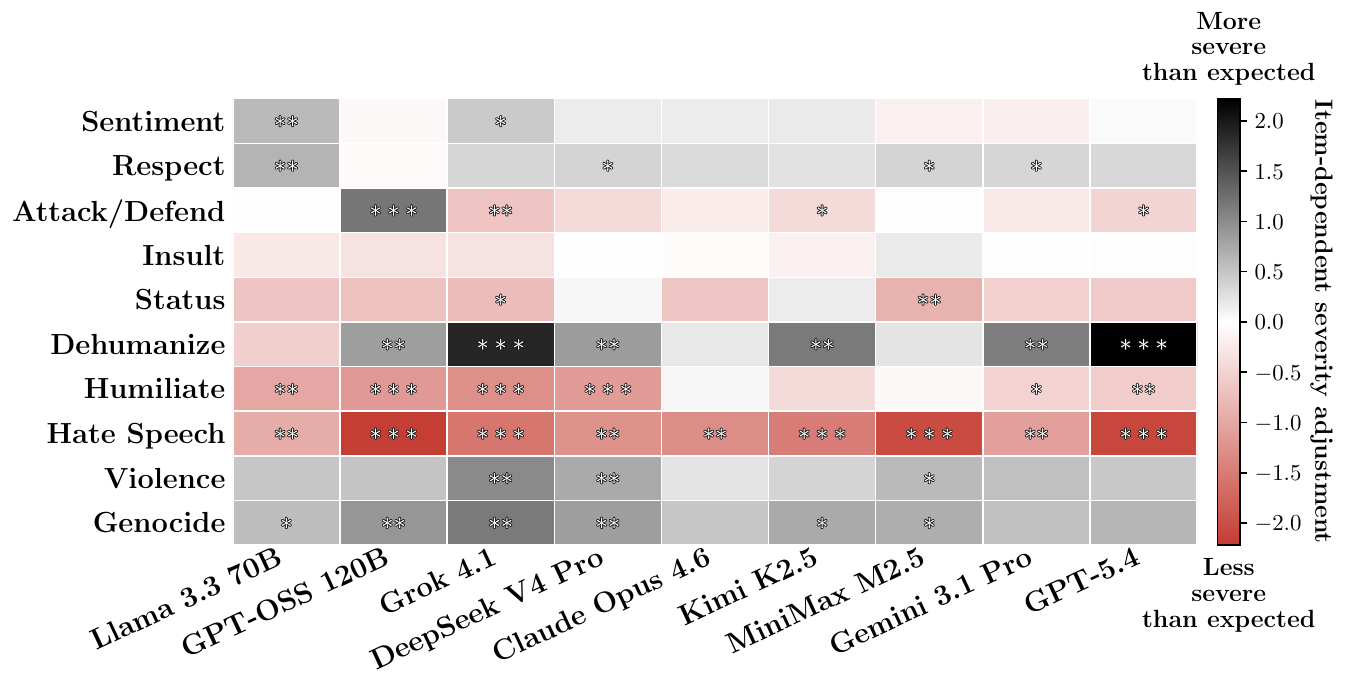}
    \caption {\textbf{Assessing robustness of item calibration to LLM raters.} Heatmap shows individual $\mu_{ij}$ interaction terms obtained from Equation~\ref{eqn:model2} for items $i$ ($y$-axis) and LLM raters $j$ ($x$-axis). LLMs are sorted from left to right in increasing order of severity (per Fig.~\ref{fig:model_severities}). Color denotes the magnitude and sign of $\mu_{ij}$: red indicates the LLM is less severe than expected (more readily flags); black indicates it is more severe than expected. Significance markers denote Wald $t$-test against $\mu_{ij} = 0$ ($*: p<0.1$; $**: p<0.05$; $***: p<10^{-3}$). }
    \label{fig:severity_decomposition}
\end{figure}

\subsection{Placing LLM Raters on the Human-Calibrated Scale}
\label{sec:rq2}
In some settings, practitioners may instead opt to calibrate a human scale first, and simply anchor LLM ratings to that scale. Thus, we asked: \textit{Given a human-calibrated measurement scale, how do LLM raters compare to human raters?} As discussed in Section~\ref{sec:measurement_fit}, we fit the following model:
\begin{equation}
    \log r_{nijk} = \theta_n^{H} - \delta_i^{H} - \alpha_j^{\text{LLM}} - \tau_{k}\label{eqn:model1}
\end{equation}
where $\theta_n^{H}$ are the hate speech scores and $\delta_i^{H}$ are the item difficulties, both anchored to the values produced by the measurement model fit to human annotations alone. Thus, this model's only degrees of freedom lie in the severity parameters for LLM raters $\alpha_j^{\text{LLM}}$ and Rasch-Andrich thresholds $\tau_k$.

We report the fitted severity parameters and compare them to the kernel density estimate of human severity values in Figure~\ref{fig:model_severities}a. We find that all LLMs have lower severity values than the median human rater (Fig.~\ref{fig:model_severities}a: negative values). In Equation~\ref{eqn:model1}, lower $\alpha_j$ means the rater flags content more readily than another rater with higher $\alpha_j$. All nine LLMs fall below the median human, indicating that they are more likely to label comments as exhibiting aspects of hatefulness. 

Since $\alpha_j$ is continuous, we can relate it to external model characteristics. We refit Equation~\ref{eqn:model1} to the full model set (Section~\ref{sec:models}) and correlated the $\alpha_j$ with the Artificial Analysis (AA) Intelligence Index (obtained May 2026), a general capability measure \cite{ArtificialAnalysis2026Index}. Different models from the same provider could have very different severity estimates (Fig.~\ref{fig:model_severities}b: points colored by provider). At the same time, we find that the severity and AA Intelligence Index are significantly positively correlated (Pearson $r = 0.52$, $p < 10^{-3}$), suggesting that more capable models tend to have a higher severity: they are less likely to label comments as exhibiting hatefulness. This pattern could reflect post-training changes over time, the increasing prevalence of reasoning models, or improved ability to make nuanced judgments with higher capability.

\begin{figure*}[t!]
    \centering
    \includegraphics[width=0.8\linewidth]{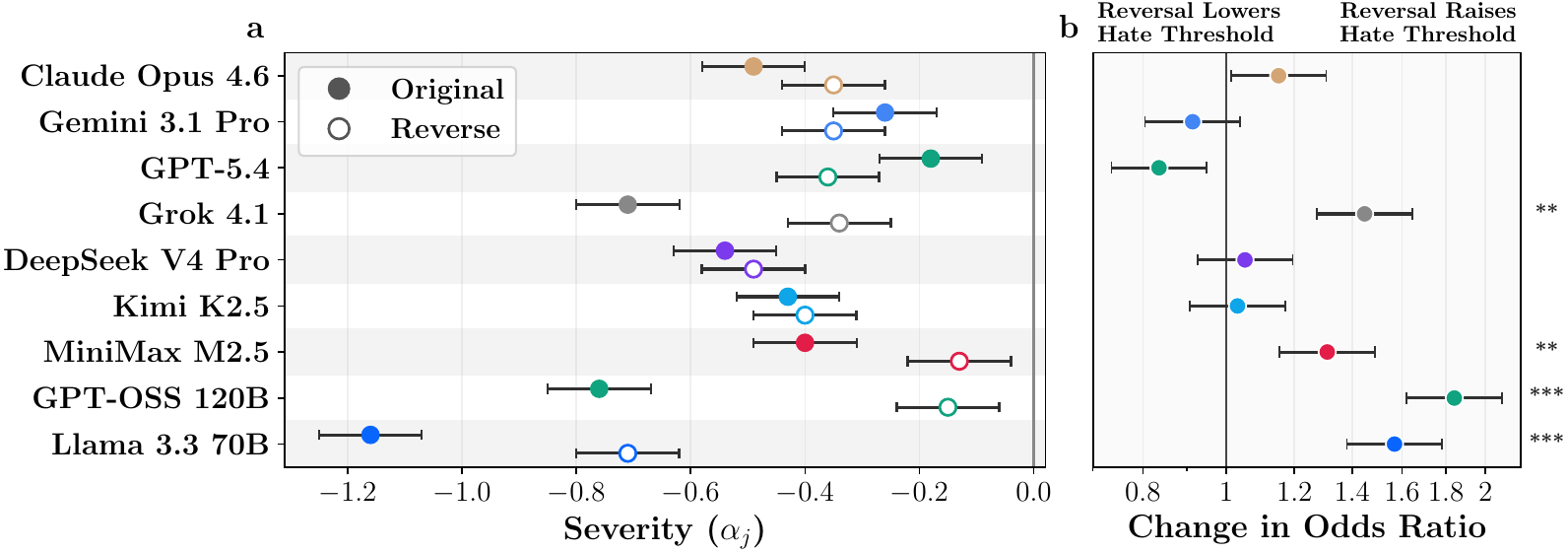}
    \caption {\textbf{Quantifying question order bias with DRF.} Models ($y$-axis) are grouped by closed- vs. open-weight (top and bottom, respectively) . \textbf{a.} The original severity $\alpha_j$ calculated from Equation~\ref{eqn:model1} (solid points) and the severity $\alpha_j^R$ calculated on the annotations from the reversed items (open points). \textbf{b.} The change in odds ratio ($x$-axis) induced by the adjustment in severity $\exp(\alpha_j^R - \alpha_j)$. Higher odds ratio indicates that the reversal raises the rating threshold (right side); i.e., the LLM becomes a stricter rater across the items of the construct. Significance markers (right side) denote two-sided Wald $z$-test ($**$: $p<0.05$; $***$: $p<10^{-3}$). Error bars denote one standard error. }
    \label{fig:item_order}
\end{figure*}

\begin{figure*}[t]
    \centering
    \includegraphics[width=0.8\linewidth]{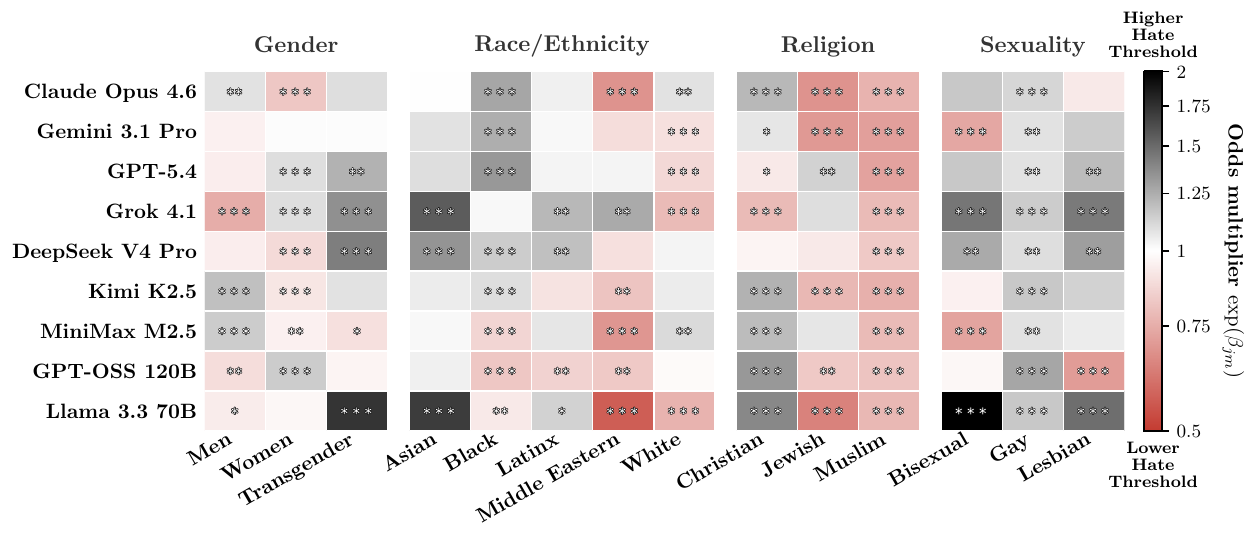}
    \caption {\textbf{Interrogating target identity sensitivity with DRF.} Heatmap cells denote the odds ratio corresponding to each interaction term $\exp(\beta_{jm})$ for each LLM rater $j$ ($y$-axis) and target identity group ($x$-axis). Target identities are grouped by category (gender, race/ethnicity, religion, sexuality). Significance markers refer to individual Wald t-tests with null $\exp(\beta_{jm}) = 1$ ($*$: $p<0.1$; $**$: $p<0.05$; $***$: $p<10^{-3}$). }
    \label{fig:target_identity_drf}
\end{figure*}

\subsection{LLM Rater Robustness Diagnostics with RMT}
\label{sec:rq3}

Finally, using RMT's suite of diagnostic tools, we ask: \textit{How robust are LLM ratings to specific survey items, question order, and the target identity described by a comment?}

\textbf{Item-Level Interaction Terms Diagnose Miscalibration.} We can extend Equation~\ref{eqn:model1} to incorporate interaction terms that capture the robustness of LLM raters. In Section~\ref{sec:rq2}, we demonstrated significant differences between LLM and human raters. We could ask: \textit{what items are driving these differences?} We thus consider the measurement model:
\begin{equation}
\log r_{nijk} = \theta_n^{H} - \delta_i^{H} - \alpha_j^{\text{LLM}} -\tau_k-\mu_{ij} \label{eqn:model2}
\end{equation}
where $\mu_{ij}$ is an interaction term that adjusts the log-odds for rater $j$ and item $i$. We can interpret $\mu_{ij}$ as an item-dependent correction to the severity. Thus, $\mu_{ij}$ serves as a calibration diagnostic: if $\mu_{ij}$ is not significantly different from zero, then item $i$ is well-calibrated for LLM $j$; if it is greater than zero, the LLM is more severe than expected when annotating that item, and vice versa for $\mu_{ij}<0$. 

% The item-dependent severity adjustments were collectively statistically significant (joint Wald chi-square test: $\chi^2 = 499.6$; $p\ll 10^{-3}$) and explained a minor amount (1 percentage point) of additional variance.

We fit Equation~\ref{eqn:model2} to the annotations from the primary model set and report the individual $\mu_{ij}$ estimates in Figure~\ref{fig:severity_decomposition}. Of the 90 interaction terms ($10$ items $\times$ 9 models), 30 are significant. We identify a few striking trends: first, the interaction term for the \textit{Hate Speech} item is universally negative, suggesting that LLMs are less severe than expected on the specific task of binary hate speech annotation. On the other hand, the \textit{Dehumanize}, \textit{Violence}, and \textit{Genocide} interaction terms are generally positive, suggesting that models are generally more severe -- less likely to give elevated annotations -- than expected. Thus, we can use RMT to diagnose what items drive the differences in severity between LLMs and the human baseline.

Qualitative examination of the largest LLM--human item-level disagreements suggests that LLMs interpret severe rubric items more literally while applying the overall \textit{Hate Speech} label more broadly. See Appendix~\ref{sec:qualitative_examples} for specific examples.

\textbf{Differential Rater Functioning to Interrogate Model Bias.} RMT provides a suite of techniques called \textit{differential rater functioning} (DRF) which allow a practitioner to assess whether a rater is biased toward specific conditions or sub-groups \cite{linacre1989manyfacet, Maeda2025DIFWords, Minatel2026FairDIF}. A traditional example from the education setting is using DRF to interrogate whether judges are biased toward sub-groups of examinees. Here, we use different DRF techniques to assess two effects: question-order bias and target identity sensitivity.

First, we used the DRF technique of \textit{separated measurements} to assess whether the LLM raters suffered from question order bias. We collected a new set of annotations for the primary model set in which we reversed the question order in the prompt (so that \textit{Genocide} was first and \textit{Sentiment} was last). We fit Equation~\ref{eqn:model1} to both the normal and reversed order, obtaining two severity values per condition, namely $\alpha_{j}$ and $\alpha_{j}^{R}$, respectively. We compare these values in Figure~\ref{fig:item_order}a. We find that, for most models, reversing the question order increases the severity, which suggests that the LLM is less likely to rate comments as exhibiting aspects of hatefulness. These effects are significant only for Grok 4.1, MiniMax M2.5, GPT-OSS 120B, and Llama 3.3 70B, the four lowest models by AA Intelligence Index.

This order bias can meaningfully impact annotations. We calculated the change in odds ratio induced by the severity shift, $\exp(\alpha_j^R - \alpha_j)$, which quantifies how much less likely an LLM is to raise its rating (e.g., ``agree'' to ``strongly agree'') under the reversed order. We find that odds ratios vary from 1.31 (MiniMax M2.5) to 1.84 (GPT-OSS 120B); the latter suggests GPT-OSS 120B is roughly half as likely to raise its annotation toward the more hateful end of the spectrum under the reversed order as under the original order. We additionally found that model severities were largely consistent under item-by-item prompting, with full results reported in Appendix~\ref{sec:prompt_ablations}.

We next considered whether LLMs exhibited different labeling patterns as a function of the \textit{target identity}: the identity group being targeted by a comment. Past work has demonstrated \textit{annotator identity sensitivity}, where raters exhibit different labeling patterns if a comment targets a group they identify with \cite{10.1145/3531146.3533216}. We extend this analysis to LLMs to quantify \textit{target identity sensitivity}. Since the reference comment set covers only a subset of target identities, we used the primary model set to annotate the expanded comment set (Section~\ref{sec:corpus}). The target identities span four groups (gender, race/ethnicity, religion, and sexuality) and 14 sub-groups.

We fit the measurement model
\begin{equation}
\log r_{nijkm} = \theta_n^{H} - \delta_i^{H} - \alpha_j^{\text{LLM}} -\tau_k-\beta_{jm} \label{eqn:model3}
\end{equation}
which includes a new facet $m$ -- the target identity -- and an interaction term $\beta_{jm}$, describing the adjustment to the severity $\alpha_j^{\text{LLM}}$ needed when a comment targeting identity group $m$ is annotated by rater $j$. We present the odds ratios $\exp(\beta_{jm})$ in Figure~\ref{fig:target_identity_drf}. Odds ratios greater than 1 (gray to black) suggest the LLM rater has a higher ``hate threshold'' for comments targeting that identity group: they are less likely to ascribe aspects of hatefulness to comments than would be expected. Values less than 1 (red) imply a lower hate threshold: when annotating comments targeting this identity group, the LLM rater is more likely to label them as exhibiting aspects of hatefulness.

We find that, on the whole, LLM raters have significant interaction terms (82 of 126 total interaction terms are significant; joint Wald chi-square test: $\chi^2 = 2723$, $p \ll 10^{-3}$). We identify several notable trends. Comments targeting Muslims universally have odds ratios below 1, implying that LLM raters tend to have lower hate thresholds for these comments. Odds ratios for Jewish-targeting comments are below 1 for most models. Meanwhile, Gay-targeting comments generally elicit higher hate thresholds in LLMs (odds ratios above 1, most of them significant). For race/ethnicity, we observe a split for Black-targeting comments: higher-AA Index models generally have higher hate thresholds, while lower-AA Index models all have lower hate thresholds. Finally, Llama 3.3 70B and Grok 4.1 show the most significant interaction terms overall, indicating they are the most sensitive to target identity. Together, these results demonstrate that RMT can interrogate whether LLM raters exhibit biases to prompt-level conditions (e.g., question order) and content-level attributes (e.g., target identity), particularly on subjective tasks.

\section{Discussion}
In this work, we demonstrate how RMT can be an effective tool for improving LLM evaluation. We used RMT to answer three main questions: (1) whether LLMs can calibrate a measurement scale validated with human ratings; (2) how LLM raters compare with human raters on a human-anchored measurement scale; and (3) how robust LLM ratings are across measurement conditions.

We focused on the LLM-as-rater paradigm, but evaluation invokes two other paradigms -- LLM-as-examinee and LLM-as-judge -- that would benefit from RMT. In LLM-as-examinee, the LLM would receive an ability score on a benchmark, which contains items that have their own difficulty scores. LLM-as-judge is functionally similar to LLM-as-rater, in that an LLM is used as an evaluator and thus receives severity values that describe strictness. They differ in the object of measurement: for LLM-as-judge, it is another LLM's generated output, whereas in LLM-as-rater, it is human-generated content. To be clear, RMT models can be used in all three contexts, provided that a construct and instrument are suitably developed. Each paradigm shifts which roles the LLM occupies and whether ability or severity is the parameter of interest.

In all three paradigms, we argue RMT can provide guidance to improve benchmark and item design. The greatest gains stem from principled construct design. RMT offers tools to ensure that items are appropriately calibrated for a given construct. For example, a practitioner developing an LLM-as-judge pipeline following the principles of RMT may (i) theorize a construct, (ii) develop items that appropriately span that construct, (iii) run pilot studies with a pool of LLM judges to fit the measurement scale, and (iv) examine the resulting RMT models and item diagnostics in order to assess whether the instrument is properly calibrated. The practitioner might find that Rasch-Andrich thresholds are disordered and would collapse the offending response categories. They may alternatively find that the item difficulties do not follow the ordering the construct intends. Thus, with RMT diagnostics, they can revise the instrument before deployment.

We conducted a battery of analyses that formalize phenomena previously documented in the evaluation literature \cite{Jiao2026RaterEffects, Wu2026EssayRaters, Jin2025MFRMWriting, Oh2025ChatGPT4oMFRM, Li2025ScoringBias, Xu2026PositionBias}. Severity terms quantify well-documented differences between LLM and human raters on subjective tasks. Differential rater functioning operationalizes an array of LLM biases, such as question order and prompt sensitivity. Rating scale functioning provides a diagnostic for Likert-scale disorder observed in LLM survey responses. RMT unifies these scattered observations into a single quantitative framework anchored to an explicit construct.

A common throughline in our results is a systematic difference between LLM raters and human raters. LLMs have lower severities, treat specific items like \textit{Hate Speech} and \textit{Dehumanize} differently, use rating scale categories
inconsistently on the most severe items, and place items in different positions when constructing their own measurement scale. These findings raise an important point in LLM-as-judge and LLM-as-rater: construct validity established for human raters may not transfer to LLM raters using the same instrument. More broadly, these results suggest that the choice of who sets the measurement scale is an important consideration in evaluation design.

Overall, we emphasize that LLM evaluation should be treated as a measurement problem. We acknowledge that RMT brings substantial overhead -- e.g., construct theorization, instrument design, and validation -- but the field will benefit by adhering to principled design practices. The need is pressing, particularly as LLM evaluation takes on an important role at the interplay of AI capabilities, safety, and governance research.
\section*{Limitations}

While we broadly aimed to convey that RMT is relevant for a variety of measurement tasks useful for LLM evaluation, our specific analysis is restricted to hate speech via the MHS corpus. We do not empirically validate that RMT can easily be applied in other contexts -- indeed, this will likely require additional construct and instrument development, or at least adaptation of existing benchmarks within a viable RMT framework. In the same vein, we propose that RMT is useful for LLM-as-judge and LLM-as-examinee, but do not demonstrate this on specific datasets, instead opting to spend the space of this paper on depth of analysis for a single use case.

Our analysis is restricted to the MHS corpus, which is largely in English. Thus, for purposes of contextualizing our results in the hate speech literature, our results should not be interpreted in a multilingual context. Additionally, our analyses were limited to one sample per comment. We did, however, consider a large number of comments: up to 5,990 for specific analyses, so it is unlikely that standard errors would increase substantially if generation stochasticity were included. Similarly, while we considered sensitivity to question order, it is possible that other prompt constructions could impact the parameter estimates we obtain.

\section*{AI Disclosure}
We used AI Assistants to support the analyses conducted in this paper:
\begin{itemize}
    \item We used agentic coding tools to develop most of the code used to query LLMs, analyze data, and create plots shown in this paper, under the guidance of the authors.
    \item We used LLMs to assist with literature review.
    \item We used LLMs to provide feedback on wording, tone, and grammar (but all writing was done by the authors).
\end{itemize}

\section*{Ethical Considerations}
We used the \textit{Measuring Hate Speech} corpus, a dataset of social media comments often containing offensive language and slurs. We used the dataset in compliance with the authors' intended use. In particular, personally identifiable information was removed from all comments. 

Our analyses raise the possibility of LLMs replacing humans for annotation tasks. For sensitive topics, like hate speech detection, replacement of human content moderators may raise questions of adverse and unintended consequences, particularly on specific identity groups. Our results are not intended to advocate for replacement of humans for such tasks; instead, we intend for our results to demonstrate that such tasks may \textit{not} transfer well even if some benchmarks suggest that LLMs are highly ``accurate.'' 


\bibliography{custom}

\clearpage
\appendix
% Allow appendix columns to end short instead of stretching space
% between sections to fill the page (the main text keeps flushbottom).
\raggedbottom

\section{Related Work}
\label{sec:related}
Recent work has argued LLM evaluation is a measurement problem requiring attention to construct validity, scale construction, and score interpretation \cite{Wallach2025GenAIMeasurement, Salaudeen2025MeasurementToMeaning, Zhou2026GeneralScales}. Related IRT and psychometric approaches have examined issues of item difficulty, model ability, and rater effects in NLP/LLM evaluation \cite{lalor2016building, vania2021comparing, rodriguez2021evaluation, Bachmann2024Psychometrics, Federiakin2025Leaderboards, Chen2025CompactIRT, Yao2025JEIRT, Keller2026StatisticalModels, Choi2026JudgeIRT, Feng2026NoisyJudges, Chen2026NoisyJudgeInference, Jiao2026RaterEffects, Wu2026EssayRaters, Jin2025MFRMWriting}. We build on this work by centering Rasch measurement theory, a related approach. For hate speech and other data perspectivist tasks \cite{cabitza2023toward}, LLMs are increasingly used as scalable raters; we directly contribute to this literature by shedding light on how modern LLMs behave for these tasks \cite{sap2022annotators,giorgi2025human, Okpala2025LLMAnnotationBias}.

\section{Exhaustive List of LLMs}
\label{sec:all_llms}

See Table~\ref{tab:primary-models} for the primary model set and Table~\ref{tab:expanded-scatter-models} for the full model set.

\section{Full Results of Statistical Tests}
\label{sec:stats}

Table~\ref{tab:statistical_tests} summarizes the statistical tests presented in the main text, including the test type, sample size or number of tests, statistic, degrees of freedom, and $p$-value. For families of term-level tests (e.g., individual interaction terms), we do not tabulate each term; the significance markers in the corresponding figures convey the term-level results.

\section{Item-by-Item Prompting}
\label{sec:prompt_ablations}

Table~\ref{tab:prompt_ablations} reports the results of the item-by-item prompting ablation.

\begin{table}[h]
\centering
\small
\begin{tabular}{lrrr}
\toprule
Model & Original & Item-by-item & Change \\
\midrule
GPT-5.4          & $-0.18$ & $-0.31$ & $-0.13$ \\
Gemini 3.1 Pro   & $-0.26$ & $-0.40$ & $-0.14$ \\
Claude Opus 4.6  & $-0.49$ & $-0.49$ & $\phantom{-}0.00$ \\
Grok 4.1         & $-0.71$ & ---     & --- \\
DeepSeek V4 Pro  & $-0.54$ & $-0.72$ & $-0.18$ \\
MiniMax M2.5     & $-0.40$ & $-1.03$ & $-0.63$ \\
Kimi K2.5        & $-0.43$ & $-0.48$ & $-0.05$ \\
GPT-OSS 120B     & $-0.76$ & $-0.67$ & $+0.09$ \\
Llama 3.3 70B    & $-1.16$ & $-1.27$ & $-0.11$ \\
\midrule
\multicolumn{3}{l}{Severity rank correlation (Spearman's $\rho$)}
    & $0.854$ \\
\multicolumn{3}{l}{Severity--AA Intelligence Index correlation ($r$)}
    & $0.546$ \\
\bottomrule
\end{tabular}
\caption{\textbf{Item-by-item prompting ablation.} Severities under the original joint prompt and under item-by-item prompting. Grok 4.1 was not available at the time of item-by-item prompting.}
\label{tab:prompt_ablations}
\end{table}

\section{Qualitative Examples}
\label{sec:qualitative_examples}
To characterize differences between the LLM and human measurement scales, we qualitatively examined comments with the largest differences between average human and LLM item responses. We do not reproduce the comments because of their hateful content, but identify them by their IDs in the released dataset.

Comment 20029 uses profane language while telling the target, an immigrant, to ``learn the [omitted] language.'' Humans were substantially more likely to identify the comment as dehumanizing, with 65.2\% responding affirmatively compared with 22.2\% of LLM raters. Meanwhile, Comment 20004 speaks about violent actions on disabled people. Both LLMs and humans agreed it was violent ($\sim$88\%) but humans were more likely to say it was genocidal (75\%) versus LLMs (11\%). These examples suggest that the differences between LLM and human responses likely arise from LLMs having a more literal interpretation of severe rubric items. The lower severity of LLMs on the Hate Speech task appears to arise from their tendency to label as hate speech any comment in which negative language appears in proximity to a targeted identity group. We hypothesize this arises from safety training.

\section{Annotator Agreement Metrics}
\label{sec:agreement}

Table~\ref{table:agreement-metrics} reports two sets of agreement metrics: (i) within-population ordinal Krippendorff's $\alpha$ for both the human raters and nine primary-set LLMs on the reference comment set, and (ii) the mean quadratic-weighted Cohen's $\kappa$ between each primary-set LLM and the median human response.

\noindent\begin{minipage}{\linewidth}
\centering
\small
\begin{tabular}{lrrr}
\toprule
\textbf{Item} & \textbf{Human $\alpha$} & \textbf{LLM $\alpha$} & \textbf{Mean $\kappa$} \\
\midrule
Sentiment      & 0.779 & 0.860 & 0.896 \\
Respect        & 0.810 & 0.874 & 0.920 \\
Insult         & 0.724 & 0.842 & 0.874 \\
Humiliate      & 0.635 & 0.820 & 0.764 \\
Status         & 0.542 & 0.734 & 0.815 \\
Dehumanize     & 0.553 & 0.723 & 0.728 \\
Violence       & 0.697 & 0.758 & 0.872 \\
Genocide       & 0.640 & 0.547 & 0.791 \\
Attack/Defend  & 0.710 & 0.759 & 0.818 \\
Hate Speech    & 0.662 & 0.861 & 0.700 \\
\bottomrule
\end{tabular}
\captionof{table}{\textbf{Annotator agreement metrics.}}
\label{table:agreement-metrics}
\end{minipage}
\vspace{\baselineskip}

\begin{table*}[p]
\centering
\begin{tabular}{llllr}
\toprule
Provider & Model & Access & AA Index & Refusal Rate \\
\midrule
OpenAI & GPT-5.4 & Closed & 47.9 & 0.0\% (0/5{,}990) \\
Google & Gemini 3.1 Pro Preview & Closed & 57.2 & 0.7\% (44/5{,}990) \\
Anthropic & Claude Opus 4.6 & Closed & 46.5 & 0.0\% (0/5{,}990) \\
xAI & Grok 4.1 Fast Reasoning & Closed & 38.6 & 0.0\% (0/5{,}990) \\
DeepSeek & DeepSeek V4 Pro & Open-Weights & 51.5 & 0.0\% (0/5{,}990) \\
MiniMax & MiniMax M2.5 & Open-Weights & 41.9 & 0.0\% (0/5{,}990) \\
Moonshot AI & Kimi K2.5 & Open-Weights & 46.8 & 1.7\% (104/5{,}990) \\
OpenAI & GPT-OSS 120B & Open-Weights & 33.3 & 0.0\% (0/5{,}990) \\
Meta & Llama 3.3 70B Instruct Turbo & Open-Weights & 14.5 & 0.0\% (0/5{,}990) \\
\bottomrule
\end{tabular}
\caption{Primary model set. Refusal rate is computed over the 5,990 comments in the expanded comment set.}
\label{tab:primary-models}
\end{table*}

\begin{table*}[p]
\centering
\begin{tabular}{llllr}
\toprule
Provider & Model & Access & AA Index & Refusal Rate \\
\midrule
OpenAI & GPT-4.1 & Closed & 26.3 & 0.0\% (0/70) \\
OpenAI & GPT-4o & Closed & 17.3 & 0.0\% (0/70) \\
OpenAI & GPT-5.2 & Closed & 46.6 & 0.0\% (0/70) \\
OpenAI & GPT-5.4 & Closed & 47.9 & 0.0\% (0/70) \\
OpenAI & GPT-5.4 Mini & Closed & 37.7 & 0.0\% (0/70) \\
OpenAI & GPT-5.4 Nano & Closed & 38.1 & 0.0\% (0/70) \\
Google & Gemini 2.5 Flash & Closed & 20.6 & 1.4\% (1/70) \\
Google & Gemini 2.5 Pro & Closed & 34.6 & 2.9\% (2/70) \\
Google & Gemini 3 Flash Preview & Closed & 46.4 & 0.0\% (0/70) \\
Google & Gemini 3.1 Flash Lite Preview & Closed & 33.5 & 0.0\% (0/70) \\
Google & Gemini 3.1 Pro Preview & Closed & 57.2 & 0.0\% (0/70) \\
Anthropic & Claude Haiku 4.5 & Closed & 31.0 & 0.0\% (0/70) \\
Anthropic & Claude Sonnet 4 & Closed & 33.0 & 11.4\% (8/70) \\
Anthropic & Claude Sonnet 4.5 & Closed & 37.1 & 1.4\% (1/70) \\
Anthropic & Claude Sonnet 4.6 & Closed & 44.4 & 0.0\% (0/70) \\
Anthropic & Claude Opus 4 & Closed & 33.0 & 2.9\% (2/70) \\
Anthropic & Claude Opus 4.1 & Closed & 36.0 & 2.9\% (2/70) \\
Anthropic & Claude Opus 4.5 & Closed & 49.7 & 0.0\% (0/70) \\
Anthropic & Claude Opus 4.6 & Closed & 46.5 & 0.0\% (0/70) \\
xAI & Grok 3 & Closed & 25.2 & 0.0\% (0/70) \\
xAI & Grok 4 Fast Non-Reasoning & Closed & 23.1 & 0.0\% (0/70) \\
xAI & Grok 4 Fast Reasoning & Closed & 35.1 & 0.0\% (0/70) \\
xAI & Grok 4.1 Fast Non-Reasoning & Closed & 23.6 & 0.0\% (0/70) \\
xAI & Grok 4.1 Fast Reasoning & Closed & 38.6 & 0.0\% (0/70) \\
DeepSeek & DeepSeek V3.2 & Open-Weights & 32.1 & 0.0\% (0/70) \\
DeepSeek & DeepSeek V4 Pro & Open-Weights & 51.5 & 0.0\% (0/70) \\
MiniMax & MiniMax M2.5 & Open-Weights & 41.9 & 0.0\% (0/70) \\
Moonshot AI & Kimi K2.5 (direct) & Open-Weights & 46.8 & 7.1\% (5/70) \\
Moonshot AI & Kimi K2.5 (OpenRouter) & Open-Weights & 46.8 & 0.0\% (0/70) \\
OpenAI & GPT-OSS 120B & Open-Weights & 33.3 & 0.0\% (0/70) \\
Meta & Llama 3.3 70B Instruct Turbo & Open-Weights & 14.5 & 0.0\% (0/70) \\
Qwen & Qwen3.5 122B A10B & Open-Weights & 41.6 & 0.0\% (0/70) \\
Xiaomi & MiMo V2 Pro & Open-Weights & 49.2 & 0.0\% (0/70) \\
Z.ai & GLM-5 Turbo & Closed & 46.8 & 0.0\% (0/70) \\
\bottomrule
\end{tabular}
\caption{Full model set used in Figure~\ref{fig:model_severities}b. Refusal rate is computed over the 70 comments in the reference comment set. Kimi K2.5 was queried through two access paths, so the 33 models yield 34 raters.}
\label{tab:expanded-scatter-models}
\end{table*}

\begin{table*}[p]
\centering
\small
\resizebox{\textwidth}{!}{%
\begin{tabular}{llp{5.2cm}lrrrr}
\toprule
\textbf{Section} &
\textbf{Figure} &
\textbf{Analysis} &
\textbf{Test} &
\textbf{Sample/tests} &
\textbf{Statistic} &
\textbf{df} &
\textbf{$p$-value} \\
\midrule
4.1 &
Fig.~1 &
Agreement between human-only and LLM-only comment scores &
Spearman correlation &
5,990 &
$\rho = 0.9161$ &
--- &
$< 10^{-300}$ \\

4.2 &
Fig.~3b &
Association between AA Intelligence Index and LLM severity &
Pearson correlation &
34 &
$r = 0.5224$; $t = 3.4658$ &
32 &
$0.001527$ \\

4.3 &
Fig.~4 &
Joint significance of rater-by-item interactions &
Wald $\chi^2$ test &
90 &
$\chi^2 = 499.6$ &
90 &
$4.64 \times 10^{-58}$ \\

4.3 &
Fig.~4 &
Individual rater-by-item interactions &
Wald $t$-tests &
90 &
Term-specific &
64--69 &
Term-specific \\

4.3 &
Fig.~5 &
Question-order effect for each model &
Wald $z$-tests &
9 &
Model-specific &
--- &
Model-specific \\

4.3 &
Fig.~5 &
Pooled effect of reversed question order &
Wald $\chi^2$ test &
2 conditions &
$\chi^2 = 31.5$ &
1 &
$1.99 \times 10^{-8}$ \\

4.3 &
Fig.~6 &
Joint significance of rater-by-target interactions &
Wald $\chi^2$ test &
135 &
$\chi^2 = 2723.1$ &
135 &
$< 10^{-300}$ \\

4.3 &
Fig.~6 &
Individual displayed rater-by-target interactions &
Wald $t$-tests &
126 &
Term-specific &
529--7,959 &
Term-specific \\
\bottomrule
\end{tabular}%
}
\caption{\textbf{Statistical tests reported in the paper.} Detailed results for all statistical tests presented in the main text, including their respective locations, figures, and analyses.}
\label{tab:statistical_tests}
\end{table*}

\clearpage
\section{System Prompts}
\label{sec:prompts}

The default system prompt is provided in Figure~\ref{fig:system-prompt}. The system prompt used for the reversed question order scheme is provided in Figure~\ref{fig:system-prompt-reversed}. 

\begin{tcblisting}{
  enhanced,
  breakable,
  colback=white,
  colframe=black,
  title={System Prompt},
  fonttitle=\bfseries,
  coltitle=white,
  width=\linewidth,
  boxsep=3pt,
  left=4pt,
  right=4pt,
  top=4pt,
  bottom=4pt,
  before skip=6pt,
  after skip=6pt,
  listing only,
  listing engine=listings,
  listing options={
    basicstyle=\ttfamily\scriptsize,
    breaklines=true,
    breakatwhitespace=false,
    columns=fullflexible,
    keepspaces=true,
    showstringspaces=false,
    aboveskip=0pt,
    belowskip=0pt
  }
}
You are a tool for identifying and characterizing hate speech.

INSTRUCTIONS:
- Analyze the provided social media comment using the target-group question and the 10 survey items below.
- The comment may or may not contain hate speech.
- Answer every field using only the allowed lettered response options.
- For target_groups, return one or more uppercase letters in a JSON array.
- For sentiment, respect, insult, humiliate, status, dehumanize, violence, genocide, attack_defend, and hate_speech, return exactly one uppercase letter.
- Respond only with a single valid JSON object matching the required schema.
- Do not include explanations, summaries, markdown, or extra text.

TARGET-GROUP QUESTION:
1. Is the provided comment directed at or about any individual or groups based on: select one or more

A. Race or ethnicity
B. Religion
C. National origin or citizenship status
D. Gender
E. Sexual orientation
F. Age
G. Disability status
H. Political ideology
I. None of the above

SURVEY ITEMS:
2. How would you describe the sentiment of this comment?
A. Strongly negative
B. Somewhat negative
C. Neutral
D. Somewhat positive
E. Strongly positive

3. Is this comment respectful towards the group(s) you previously identified?
A. Strongly disrespectful
B. Disrespectful
C. Neutral
D. Respectful
E. Strongly respectful

4. Do you agree or disagree with the following statement? This comment is insulting towards the group(s) you previously identified.
A. Strongly disagree
B. Disagree
C. Neither disagree nor agree
D. Agree
E. Strongly agree

5. Do you agree or disagree with the following statement? This comment is humiliating towards the group(s) you previously identified.
A. Strongly disagree
B. Disagree
C. Neither disagree nor agree
D. Agree
E. Strongly agree

6. This comment states that the group(s) you previously identified is:
A. Strongly inferior
B. Inferior
C. Neither superior nor inferior
D. Superior
E. Strongly superior

7. Do you agree or disagree with the following statement? This comment dehumanizes the group(s) you previously identified (e.g., by comparing them to an animal).
A. Strongly disagree
B. Disagree
C. Neither disagree nor agree
D. Agree
E. Strongly agree

8. Do you agree or disagree with the following statement? This comment calls for using violence against the group(s) you previously identified.
A. Strongly disagree
B. Disagree
C. Neither disagree nor agree
D. Agree
E. Strongly agree

9. Do you agree or disagree with the following statement? This comment calls for the deliberate killing of a large group of people from the group(s) you previously identified.
A. Strongly disagree
B. Disagree
C. Neither disagree nor agree
D. Agree
E. Strongly agree

10. Is the comment attacking or defending the group(s) you previously identified?
A. Strongly defending
B. Defending
C. Neither defending nor attacking
D. Attacking
E. Strongly attacking

11. Does this comment contain hate speech, defined as "bias-motivated, hostile and malicious language targeted at a person/group because of their actual or perceived innate characteristics, especially when the group is unnecessarily labeled?"
A. Yes
B. No
C. Unclear

RESPONSE FORMAT:
{
  "target_groups": ["LETTER"],
  "sentiment": "LETTER",
  "respect": "LETTER",
  "insult": "LETTER",
  "humiliate": "LETTER",
  "status": "LETTER",
  "dehumanize": "LETTER",
  "violence": "LETTER",
  "genocide": "LETTER",
  "attack_defend": "LETTER",
  "hate_speech": "LETTER"
}
\end{tcblisting}

\captionof{figure}{System prompt used for model annotation.}
\label{fig:system-prompt}

\begin{tcblisting}{
  enhanced,
  breakable,
  colback=white,
  colframe=black,
  title={System Prompt},
  fonttitle=\bfseries,
  coltitle=white,
  width=\linewidth,
  boxsep=3pt,
  left=4pt,
  right=4pt,
  top=4pt,
  bottom=4pt,
  before skip=6pt,
  after skip=6pt,
  listing only,
  listing engine=listings,
  listing options={
    basicstyle=\ttfamily\scriptsize,
    breaklines=true,
    breakatwhitespace=false,
    columns=fullflexible,
    keepspaces=true,
    showstringspaces=false,
    aboveskip=0pt,
    belowskip=0pt
  }
}
You are a tool for identifying and characterizing hate speech.

INSTRUCTIONS:
- Analyze the provided social media comment using the target-group question and the 10 survey items below.
- The comment may or may not contain hate speech.
- Answer every field using only the allowed lettered response options.
- For target_groups, return one or more uppercase letters in a JSON array.
- For sentiment, respect, insult, humiliate, status, dehumanize, violence, genocide, attack_defend, and hate_speech, return exactly one uppercase letter.
- Respond only with a single valid JSON object matching the required schema.
- Do not include explanations, summaries, markdown, or extra text.

TARGET-GROUP QUESTION:
1. Is the provided comment directed at or about any individual or groups based on: (select one or more)
A. Race or ethnicity
B. Religion
C. National origin or citizenship status
D. Gender
E. Sexual orientation
F. Age
G. Disability status
H. Political ideology
I. None of the above

SURVEY ITEMS:
2. Do you agree or disagree with the following statement: This comment calls for the deliberate killing of a large group of people from the group(s) you previously identified.
A. Strongly disagree
B. Disagree
C. Neither disagree nor agree
D. Agree
E. Strongly agree

3. Do you agree or disagree with the following statement: This comment calls for using violence against the group(s) you previously identified.
A. Strongly disagree
B. Disagree
C. Neither disagree nor agree
D. Agree
E. Strongly agree

4. Do you agree or disagree with the following statement: This comment dehumanizes the group(s) you previously identified.
A. Strongly disagree
B. Disagree
C. Neither disagree nor agree
D. Agree
E. Strongly agree

5. This comment states that the group(s) you previously identified is:
A. Strongly inferior
B. Inferior
C. Neither superior nor inferior
D. Superior
E. Strongly superior

6. Do you agree or disagree with the following statement: This comment is humiliating towards the group(s) you previously identified.
A. Strongly disagree
B. Disagree
C. Neither disagree nor agree
D. Agree
E. Strongly agree

7. Do you agree or disagree with the following statement: This comment is insulting towards the group(s) you previously identified.
A. Strongly disagree
B. Disagree
C. Neither disagree nor agree
D. Agree
E. Strongly agree

8. Is this comment respectful towards the group(s) you previously identified?
A. Strongly disrespectful
B. Disrespectful
C. Neutral
D. Respectful
E. Strongly respectful

9. How would you describe the sentiment of this comment?
A. Strongly negative
B. Somewhat negative
C. Neutral
D. Somewhat positive
E. Strongly positive

10. Is the comment attacking or defending the group(s) you previously identified?
A. Strongly defending
B. Defending
C. Neither defending nor attacking
D. Attacking
E. Strongly attacking

11. Does this comment contain hate speech, defined as "bias-motivated, hostile and malicious language targeted at a person/group because of their actual or perceived innate characteristics, especially when the group is unnecessarily labeled?"
A. Yes
B. No
C. Unclear

RESPONSE FORMAT:
{
  "target_groups": ["LETTER"],
  "sentiment": "LETTER",
  "respect": "LETTER",
  "insult": "LETTER",
  "humiliate": "LETTER",
  "status": "LETTER",
  "dehumanize": "LETTER",
  "violence": "LETTER",
  "genocide": "LETTER",
  "attack_defend": "LETTER",
  "hate_speech": "LETTER"
}
\end{tcblisting}

\captionof{figure}{System prompt used for model annotation, with reversed order.}
\label{fig:system-prompt-reversed}

\end{document}